\documentclass[letterpaper, 10 pt, conference]{ieeeconf}
\usepackage{textcomp}
\usepackage{multicol}
\usepackage{multirow}

\IEEEoverridecommandlockouts
\usepackage{gensymb}
\usepackage{textgreek}
\usepackage{graphicx}
\usepackage{amsmath}
\usepackage{cite}
\usepackage{booktabs}  
\usepackage{makecell}   
\usepackage[table]{xcolor}
\usepackage{colortbl}

\usepackage{url}
\usepackage{tabularx}
\usepackage{siunitx}
\usepackage{amssymb}
\usepackage{gensymb}
\usepackage{placeins}
\begin{document}

\title{\LARGE \bf
Generating Personalized Lower-Limb Kinematics

Across Walking Speeds Using Subject-Conditioned Diffusion}

\author{\large Diya Dinesh\textsuperscript{*}, Adrian Krieger\textsuperscript{*}, Changseob Song, Dongho Park, \\
Aaron J. Young,~\IEEEmembership{Senior Member,~IEEE}, Inseung Kang,~\IEEEmembership{Member,~IEEE}
\thanks{\textsuperscript{*}These authors contributed equally to this work. (Corresponding author: Changseob Song, {\tt\footnotesize changseob@cmu.edu})}
\thanks{D. Dinesh is with the Kenneth C. Griffin School of Computer Science, Carnegie Mellon University, Pittsburgh, PA, 15213 USA. A. Krieger, C. Song and I. Kang are with the Department of Mechanical Engineering, Carnegie Mellon University, USA. D. Park and A. J. Young are with the Woodruff School of Mechanical Engineering and the Institute for Robotics and Intelligent Machines, Georgia Institute of Technology, Atlanta, GA 30332 USA.}
\thanks{This research was supported by the NIH R21 Award 1R21EB037268-01 and NIH DP2 Award 1DP2HD111709-01.}
}

\maketitle
\begin{abstract}
Personalized gait modeling supports assistive technologies, biomechanical modeling, and rehabilitation. Gait changes with walking speed, but collecting user-specific recordings across speeds is costly and burdensome, particularly for clinical populations. We propose a subject-conditioned residual diffusion model to generate target-speed gait from a single source-speed recording. A transformer conditions on source gait and source and target speeds to generate residual changes in sagittal-plane hip, knee, and ankle kinematics. Trained only on able-bodied data, the model was evaluated on cross-speed generation for held-out able-bodied and stroke subjects using kinematic error, personalization rank, cadence, and inter-joint coordination. Mean absolute kinematic errors were $3.27^\circ$ and $6.48^\circ$ for the able-bodied and stroke groups, respectively, without stroke-specific normalization or fine-tuning. The model outperformed a matched deterministic transformer, and loss ablations suggested improved stroke generalization. These results suggest potential to reduce data collection for personalized gait modeling, with applications in wearable and rehabilitation robotics.
\end{abstract}


\section{Introduction}
Personalized gait modeling is important for assistive technologies such as exoskeletons, prosthetics, and functional electrical stimulation, biomechanical models such as digital twins and user state estimators, and rehabilitation. Because gait varies substantially across individuals, walking speeds, impairments, and biomechanical constraints \cite{Winter1984, Son2009, Embry2018}, particularly in clinical populations such as stroke survivors \cite{Balaban2014stroke, Awad2017, Kang2025}, subject-specific modeling is essential for interventions that improve the user's functional outcomes such as energy \cite{Pruyn2026, Kang2025, Gunnell2025, Awad2017}, fatigue \cite{Divekar2025, Zhang2025}, and stability \cite{Awad2017, Archangeli2026, Lerner2017SciTranslMed, Kim2024}.

A major bottleneck for data-driven personalization is the large amount of user-specific gait data required across locomotor contexts \cite{Winter1984, Son2009}, since repeated motion capture sessions are time-consuming and exhausting for clinical populations. Generative models can reduce this data overhead \cite{Manduchi2024} by using gait from a single task to generate plausible data for other relevant, hard-to-collect tasks. Prior motion generation work has explored adversarial, variational, and transformer-based approaches \cite{barsoum2018hpgan, petrovich2021action, tevet2023motionclip, guo2020action2motion, lucas2022posegpt}, and diffusion models have recently shown stable training, high sample quality, and flexible conditioning for motion synthesis \cite{zhang2024motiondiffuse, Tan2026}.

Most related to this work is GaitDynamics \cite{Tan2026}, a generative foundation model that jointly models center-of-mass velocities, full-body kinematics, and ground reaction forces during walking and running. Although it can generate gait across speeds, it targets population-level modeling and was not designed to transfer a subject-specific gait pattern across locomotor contexts, which is essential for downstream wearable robot interventions.

We therefore propose a subject-conditioned residual diffusion framework that generates identity-preserving, target-speed lower-limb kinematics from a single source-speed gait sequence. Our contributions include: 1) a transformer-based denoiser conditioned on subject gait and source and target speeds; 2) a residual formulation that models kinematic changes across speeds; and 3) an evaluation demonstrating personalized gait generation on held-out able-bodied and stroke subjects without stroke-specific fine-tuning. In doing so, the framework generates valuable user-specific data from sparse task conditions, with potential to reduce the data collection burden for downstream exoskeleton personalization workflows.

\begin{figure*}[!t]
    \centering
    \includegraphics[width=\textwidth]{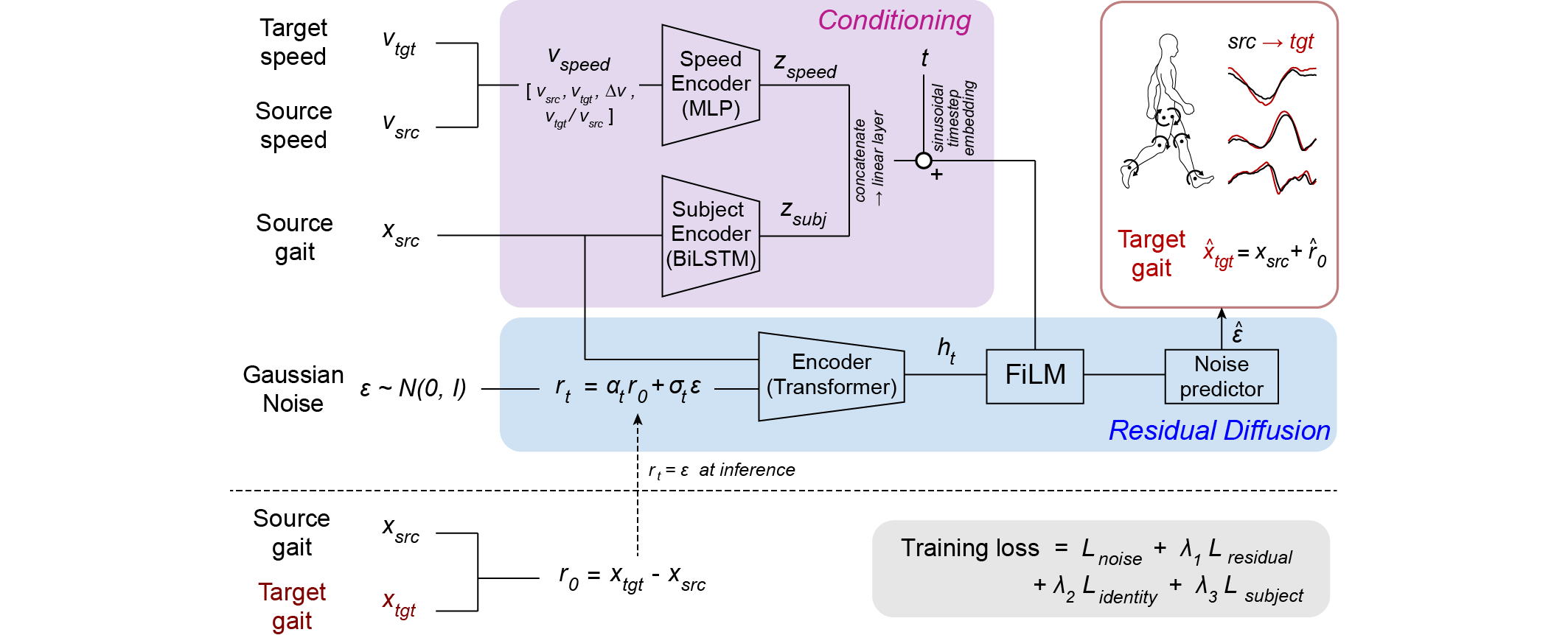}
    \caption{\textbf{Subject-conditioned residual diffusion for personalized gait synthesis.} A bidirectional long short-term memory (BiLSTM) encoder represents the subject's source gait, and a multilayer perceptron (MLP) encodes the source and target speeds. These embeddings, together with the diffusion timestep embedding, condition the transformer denoiser through feature-wise linear modulation (FiLM). The model adds the denoised residual to the source gait to generate the target-speed trajectory. Training combines noise prediction, residual reconstruction, identity preservation, and subject preservation losses.}
    \label{fig:architecture}
\end{figure*}

\section{Methods}
\subsection{Problem Design}
The core task is to generate personalized lower-limb joint kinematics for a given subject across multiple walking speeds. Given a source gait sequence $x_{src} \in \mathbb{R}^{T\times F}$ from subject $s$ walking at a source speed $v_{src}$, and a desired target speed $v_{tgt}$, the goal is to generate a gait sequence $\hat{x}_{tgt}$ at $v_{tgt}$. Here, the source sequence and the generated target sequence have the temporal dimension of $T=600$ data points, equivalent to 30 seconds at a sampling rate of 20 Hz, while the feature dimension $F=6$ captures the sagittal-plane joint angles of left and right hip, knee, and ankle joints.

The model generates source to target kinematic residuals, rather than synthesizing the target from scratch, preserving subject-specific gait patterns while capturing speed-dependent changes. The ground-truth residual is:
\begin{equation}
    r_0 = x_{tgt} - x_{src}
\end{equation}
The model then predicts the residual conditioned on the source gait and the source and target speeds:
\begin{equation}
    \hat{r}_{0} = f_{\theta}(x_{src},v_{src},v_{tgt})
\end{equation}
And finally, the target gait sequence is reconstructed as
\begin{equation}
    \hat{x}_{tgt} = x_{src}+\hat{r}_0
\end{equation}

\subsection{Gait Dataset}
The able-bodied dataset includes treadmill walking data of 22 subjects (13 males and 9 females, age 21.6 $\pm$ 3.6 years, body mass 68.5 $\pm$ 11.33 kg, height 171 $\pm$ 7 cm) across speeds from 0.5 m/s to 1.85 m/s in 0.05 m/s increments \cite{Camargo2021}. An additional stroke dataset was collected under a protocol approved by the Georgia Institute of Technology Institutional Review Board, and all participants provided written informed consent. The dataset included 19 stroke subjects (16 males and 3 females, age 56.6 $\pm$ 13.7 years, body mass 87.46 $\pm$ 14.69 kg, height 175.16 $\pm$ 6.28 cm, time since stroke 97 $\pm$ 59 months, 11 left/8 right paretic) walking on the treadmill at 0.3, 0.6, 0.9, 1.2 m/s, and a single self-selected walking speed. All data were recorded at 200 Hz with marker-based motion capture (Vicon, Oxford, UK) during unassisted treadmill walking without the participants wearing lower-limb exoskeletons. Each speed trial included at least 30 seconds of level treadmill walking with sagittal-plane lower-limb joint kinematics derived from motion capture.

\subsection{Data Preprocessing}
Each speed trial was cropped to a single 30-second segment and downsampled from 200 Hz to 20 Hz (600 time steps) to reduce the computational cost of the transformer-based diffusion model. Each feature was z-score normalized using its mean and standard deviation (SD) computed from the able-bodied training set. These statistics were applied unchanged to the validation, test, and stroke data, without stroke-specific normalization or fine-tuning. Using shared training-set statistics retained between-subject and between-speed differences in feature means and amplitudes.

After preprocessing, source and target speeds were paired within each subject across walking speeds. Training pairs were generated to randomly sample across identity/same-speed pairs and cross-speed pairs, which span the entire spectrum of possible speed differences. For the held-out validation and test subjects, pairs were constructed exhaustively from all available within-subject speed combinations.

Source and target recordings were not aligned by gait events, so the residual may include initial gait phase offsets and phase drift caused by differences in cadence, in addition to speed-dependent kinematic changes. We evaluated the generated gait using gait-cycle-normalized profiles and cadence. These measures assess kinematic similarity and walking rhythm but do not measure exact frame-level correspondence. Since our focus was on generating plausible kinematic data rather than synchronized kinematic and kinetic data, explicit gait-event alignment was not imposed in the current framework. Applications requiring synchronized trajectories would need explicit alignment or phase conditioning.

\subsection{Model Architecture}
The generative model used in this work was a conditioned residual diffusion model, which generated the residual change needed to transform the subject's source speed gait data into their target speed gait data (Fig. \ref{fig:architecture}). The architecture had three main components outlined below.

\subsubsection{Diffusion Denoiser}
At each diffusion step $t$, the model took the concatenated vector of the noisy residual sample $r_t$ and the source gait sequence $x_{src}$ as its input (size $T \times 2F$ equivalent to 12 features per timestep). The denoiser predicted the added noise $\hat{\epsilon}$ rather than the clean residual $r_0$ directly, which follows standard diffusion model training \cite{ho2020}. The predicted clean residual $\hat{r}_0$ was then added to $x_{src}$ to generate $\hat{x}_{tgt}$. The diffusion denoiser was built on the transformer encoder backbone from GaitDynamics \cite{Tan2026}.

\subsubsection{Subject and Speed Conditioning}
The diffusion denoiser was conditioned on the source gait sequence of the target user, the source-target speed parameters, and the diffusion timestep. For subject conditioning, we encoded source gait data $x_{src}$ into a lower-dimensional embedding space using a bidirectional long short-term memory (BiLSTM) encoder, and applied $L_2$ normalization to obtain a subject embedding $z_{subj}$. Because subject conditioning was derived from source kinematics rather than a subject label, the model can be applied to unseen subjects at test time. To encode the relationship between the source and target speeds, we constructed the vector
\begin{equation}
\left[ v_{src}, v_{tgt}, v_{tgt} - v_{src}, \frac{v_{tgt}}{v_{src} + 10^{-6}} \right]
\end{equation}
which was passed through a multilayer perceptron (MLP) to get the speed embedding $z_{speed}$. The small constant $10^{-6}$ in the denominator was included for numerical stability, although all source walking speeds in our dataset were strictly positive. The subject and speed embeddings were concatenated and passed through a linear layer to form a conditioning vector $c$, which captures both subject-specific gait pattern and the source-to-target mapping.

At each iteration during training, a diffusion timestep $t \in \{0, \dots, 999\}$ was sampled randomly from the 1,000 step noise schedule, encoded with sinusoidal embedding, projected by an MLP to the conditioning dimension, and added to the subject-speed conditioning vector $c$. The final conditioning vector modulated the diffusion transformer encoder's hidden states with scale and shift parameters $(\gamma, \beta)$, through feature-wise linear modulation (FiLM) \cite{perez2017filmvisualreasoninggeneral}
\begin{equation}
    \text{FiLM}(h,c) = (1+\gamma(c)) \odot h + \beta(c)
\end{equation}
where $h$ is the transformer's hidden state and $\gamma(c)$ and $\beta(c)$ are the scale and shift parameters produced from the conditioning vector $c$. FiLM modulates hidden representations shared across all six joint channels, rather than directly scaling or shifting individual joint trajectories.

\subsubsection{Noise Scheduling and Training Objective}
The model was trained using a standard variance-preserving diffusion objective on the residual gait sequence $r_0$. At each diffusion step $t$, Gaussian noise $\epsilon \sim \mathcal{N}(0,I)$ was added to the clean residual $r_0$ to obtain the noisy residual
\begin{equation}
    r_t = \alpha_t \, r_0 + \sigma_t\epsilon
\end{equation}
where $\alpha_t$ and $\sigma_t$ are determined by the variance-preserving noise schedule. The denoiser was trained to predict the added noise $\epsilon$ from $r_t$, while its intermediate hidden states were conditioned on the source gait, speed, and timestep through FiLM.

The training objective combined four loss terms, including a weighted noise prediction loss, residual reconstruction loss, identity preservation loss, and a subject preservation loss
\begin{equation}
    L=L_{noise} + \lambda_1 L_{residual}+\lambda_2 L_{identity}+\lambda_3 L_{subject}
\end{equation}
where $\lambda_1 = 0.1$, $\lambda_2 = 0.5$, and $\lambda_3 = 0.1$ are base scalar weights balancing the contribution of each loss term. The primary term is a weighted mean squared error between the predicted and true diffusion noise, where knee joints were weighted twice as heavily as hip and ankle to penalize the typical higher prediction error on estimating knee joint angle \cite{kneeAngle, Li2024}. The residual reconstruction loss compared the reconstructed clean residual to the true residual. The identity preservation loss encouraged near-zero residuals for identity mappings ($v_{src} = v_{tgt}$). The subject preservation loss compared subject encoder embeddings of the source sequence $x_{src}$ and the generated target speed sequence $\hat{x}_{tgt} = x_{src}+\hat{r}_0$ using a batch-wise contrastive loss, where the embeddings from the same subject form positive pairs and embeddings from different subjects form negative pairs, discouraging collapse into a population-average gait. The loss weights were adjusted based on the source and target speed absolute differences using piecewise-linear weighting functions. Speed differences below 0.1 m/s emphasized identity and subject preservation, while differences above 0.1 m/s progressively increased the residual reconstruction weight, reaching its maximum at 0.3 m/s.

\subsection{Training and Inference}
The able-bodied subjects were split into 13 training subjects, 4 validation subjects, and 5 test subjects, with no subject shared across splits. The main diffusion model was trained for 100 epochs, and the checkpoint with the lowest validation diffusion loss was used for all test evaluations to prioritize the goal of generalization on unseen subjects. The training used up to 100,000 balanced source-target pairs, and validation was performed using only within-subject pairs from the validation subjects. 

The model was trained using the AdamW optimizer with a learning rate of $5\times10^{-4}$, weight decay of $10^{-3}$, and batch size of 16. Mixed precision training and gradient clipping with a maximum norm of 1.0 were used for stability. We utilized a variance-preserving diffusion process with a 1,000 step noise schedule. Main hyperparameters included a conditioning dimension of 64 and a timestep embedding dimension of 128, built upon a 4-layer transformer denoiser backbone. At inference time, the model sampled from residual space using denoising diffusion implicit models sampling \cite{song2021denoising} starting from Gaussian noise. Conditioned on $x_{src}$, $v_{src}$, and $v_{tgt}$, the model denoised the sample to predict $\hat{r}_0$ over 100 steps and reconstructed the target sequence as $\hat{x}_{tgt} = x_{src}+\hat{r}_0$.

\subsection{Evaluation Metrics}
Generated gait sequences were evaluated using gait cycle-normalized mean absolute error (MAE), subject personalization rank, cadence, and inter-joint coordination. 
\subsubsection{Mean Absolute Error}
To compare generated and real gait trajectories, we computed gait cycle-normalized MAE over lower-limb joints. Gait cycles were first identified separately for the real and generated trajectories by detecting the hip extension peaks and shifted by 50\% of the detected gait cycle length to match with the nominal method based on heel-strike detection. The detected cycles were then reused for the corresponding knee and ankle joints on the same side to ensure all joints were evaluated over consistent gait-cycle boundaries. After detection, each real gait cycle was matched to a generated gait cycle greedily by selecting unused generated cycles with the closest cycle center and most similar cycle length to the real cycle. These cycles were then time-normalized to 101 points for a gait cycle percentage scale. For each joint, we averaged the normalized gait cycles for generated and real trajectories respectively, and computed the MAE between the resulting mean gait cycle trajectories for real and generated gait. The overall MAE for a single generated trajectory was then obtained by averaging the MAE values across all the lower-limb joint angles on both the right and left sides. This metric evaluates the similarity of the mean gait-cycle kinematic profiles while accounting for differences in initial gait phase between the real and generated trajectories. We calculated the Pearson correlation coefficient ($r$) between the mean gait-cycle trajectories for each joint and averaged across joints in the same way. To further assess temporal characteristics, we separately evaluated the cadence to determine how well the generated motion preserves the walking rhythm at the target speed.

\subsubsection{Personalization Metric}
To evaluate whether generated gait captures subject-specific rather than population-average features, we computed the MAE between each generated sample and the real target-speed data of every subject in the cohort with available recordings at the corresponding target speed and ranked subjects in ascending order of MAE. The personalization rank is the rank of the true subject where values closer to 1 indicate stronger subject identity preservation with the maximum rank bounded by the size of the subject cohort.

\subsubsection{Cadence}
We used cadence as a metric to evaluate how well the rhythm of the generated motion matches the target gait. Gait cycles were identified by detecting the hip extension peaks, without the $50\%$ shift used for MAE. Cadence was then computed as $2\times60/T_s$, where $T_s$ is the mean stride duration in seconds. To reduce the effect of missed peaks, strides shorter than 0.5 or longer than 1.5 times the median stride duration were excluded. Cadence was computed separately from the left and right hip trajectories and then averaged to obtain a single cadence value for the real and generated sequences.

\subsubsection{Inter-Joint Coordination} 
To evaluate inter-joint coordination, we computed the multiple coefficient of determination ($R^2$) of each joint regressed on the remaining joints using the original temporal trajectories, separately for the real and generated target gait. Bilateral lower-limb coordination used the other five joints and intra-limb coordination used the other two joints on the same leg, with stroke limbs separated into paretic and non-paretic. The absolute difference $|\Delta R^2|$ was computed for each subject and source-target speed pairing, separately for each joint, before averaging. All reported values are unweighted means over speed pairings and joints (and both limbs for able-bodied subjects), so $|\Delta R^2|$ need not equal the absolute difference between the mean $R^2$ values.

\subsection{Baseline Models}
To evaluate the diffusion approach, conditioning mechanism, and auxiliary losses, all baselines and ablations, and a copy of the full diffusion model retrained for comparison, were trained for 30 epochs using the same training, validation, and test splits. First, we implemented a deterministic transformer residual predictor using the same transformer backbone and FiLM conditioning mechanism as the proposed model, but without diffusion. This baseline was used to isolate the contribution of iterative denoising in diffusion. We also evaluated two temporal convolutional network (TCN) baselines, with FiLM and input-channel conditioning respectively, to compare a convolutional backbone against the attention-based transformer and to assess whether modulating the intermediate representations integrates subject and speed information better than concatenating it with the input. Finally, to evaluate auxiliary losses, we trained three diffusion variants by removing the identity preservation loss, subject preservation loss, and both together. All variants were evaluated excluding source-target pairs with identical walking speeds.

\section{Results}
\subsection{Overall Results}

\begin{table}[!b]
    \centering
    \caption{Overall Results on Able-Bodied and Stroke Subjects}
    \label{tab:overallMAE}
    \begin{tabular}{lcc}
    \toprule
     & Able-Bodied & Stroke \\
    \midrule
    MAE & 3.27 $\pm 0.02^{\circ}$ & 6.48 $\pm 0.09^{\circ}$ \\
    $r$ & 0.96 & 0.87 \\
    Personalization rank & 3.3 & 3.0 \\
    \bottomrule
    \end{tabular}
\end{table}

Evaluating only non-identity source-target speed pairings, the model achieved a $3.27{^\circ}$ MAE and $r$ of 0.96 on the held-out able-bodied subjects and a $6.48{^\circ}$ MAE and $r$ of 0.87 on the stroke subjects (Table \ref{tab:overallMAE}). Each non-identity test pair was generated ten times using different random seeds with the model checkpoint fixed. We computed aggregate MAE separately for each seed. The standard deviation across the 10 aggregate values was $0.02^{\circ}$ for able-bodied and $0.09^{\circ}$ for stroke subjects. The average personalization rank was 3.3 out of 22 available able-bodied subjects and 3.0 out of 19 stroke subjects. Paretic and non-paretic results are shown in Table \ref{tab:paretic}. The knee joint yielded the highest error among all joints. On the paretic side, its MAE was $2.6{^\circ}$ and $0.7{^\circ}$ larger than the hip and ankle, respectively. On the non-paretic side, the knee MAE was $2.6{^\circ}$ and $1.1{^\circ}$ larger than the hip and ankle, respectively.

\begin{figure}[!t]
    \includegraphics[width=\columnwidth]{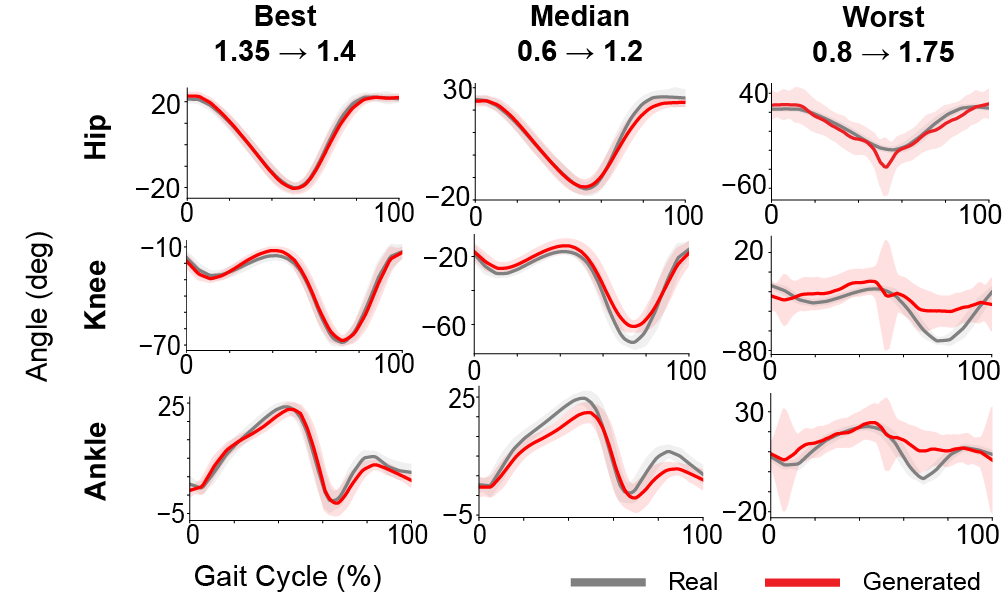}
    \caption{\textbf{Representative able-bodied gait generation.} Comparison of ground-truth and generated joint angle trajectories over a normalized gait cycle (mean$\pm$SD). Trajectories are presented for the best, median, and worst-performing speed pairings ($\text{source}\rightarrow\text{target speed}$) of the representative able-bodied subject exhibiting median overall error.}
    \label{fig:able_bodied_representative}
\end{figure}

\begin{table}[!t]
    \centering
    \caption{Kinematic Accuracy Across Stroke Subject Joints}
    \label{tab:paretic}
    \begin{tabular}{lcccc}
    \toprule
     & \multicolumn{2}{c}{MAE} & \multicolumn{2}{c}{$r$} \\
    \cmidrule(lr){2-3} \cmidrule(lr){4-5}
     & Paretic & Non-Paretic & Paretic & Non-Paretic \\
    \midrule
    Hip & 4.4${^\circ}$ & 5.7${^\circ}$ & 0.95 & 0.93 \\
    Knee & 7.0${^\circ}$ & 8.3${^\circ}$ & 0.90 & 0.90 \\
    Ankle & 6.3${^\circ}$ & 7.2${^\circ}$ & 0.78 & 0.82 \\
    Overall & 5.9${^\circ}$ & 7.1${^\circ}$ & 0.87 & 0.88 \\
    \bottomrule
    \end{tabular}
\end{table}

\begin{figure}[t]
    \centering
    \includegraphics[width=\columnwidth]{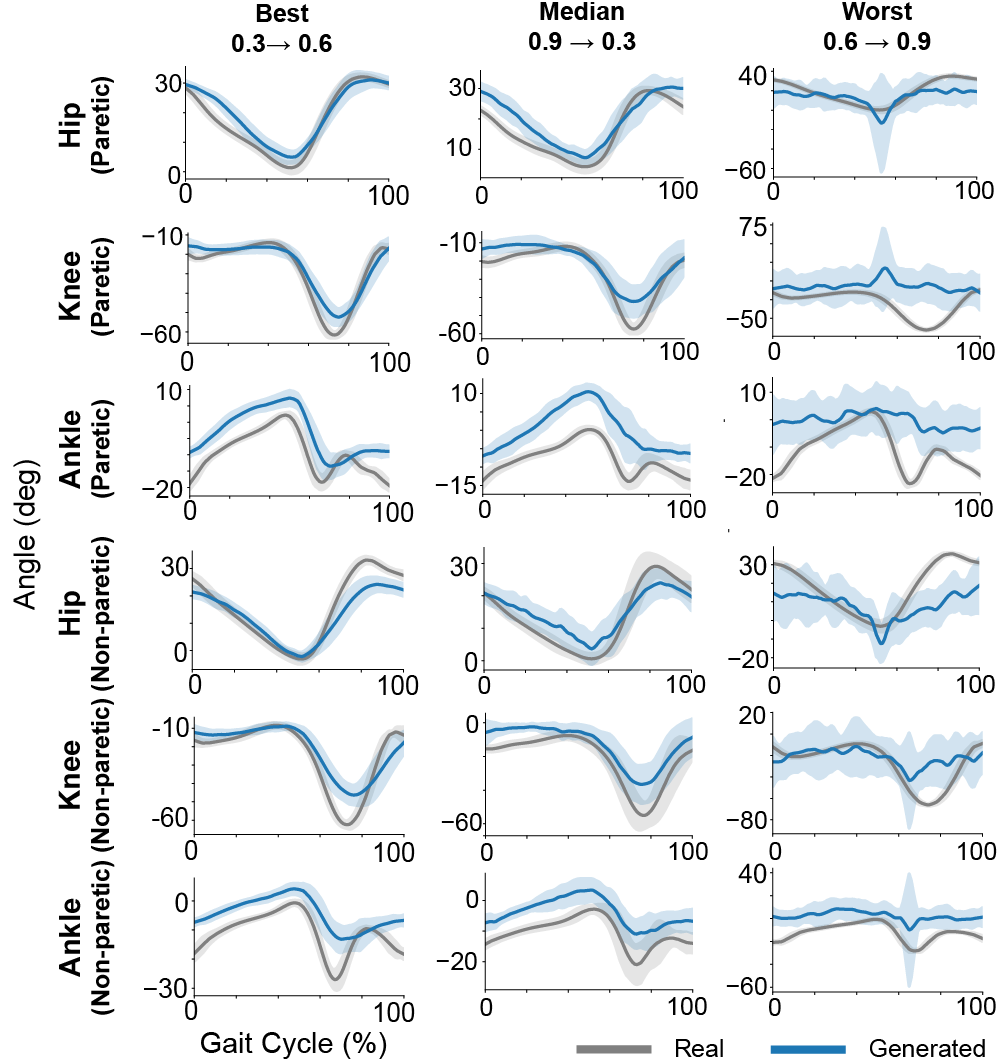}
    \caption{\textbf{Representative stroke gait generation.} Comparison of ground-truth and generated joint angle trajectories over a normalized gait cycle for both the paretic and non-paretic limbs (mean$\pm$SD). Trajectories are presented for the best, median, and worst-performing speed pairings ($\text{source}\rightarrow\text{target speed}$) of the representative stroke subject exhibiting median overall error.}
    \label{fig:stroke_representative}
\end{figure}

\begin{figure}[t]
    \includegraphics[width=\columnwidth]{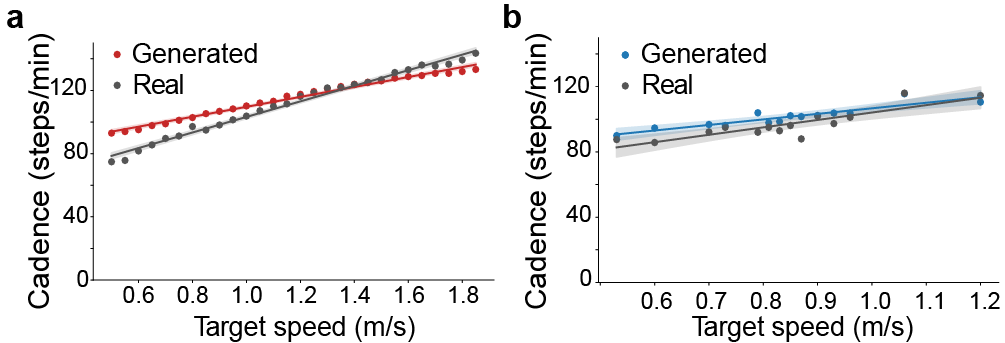}
    \caption{\textbf{Cadence of generated gait.} Each point shows mean cadence at a target speed, aggregated across source speeds and subjects. Generated and ground-truth cadence are shown alongside regression lines and 95\% confidence intervals for (a) able-bodied controls and (b) stroke individuals.}
    \label{fig:cadence}
\end{figure}

Among different generation scenarios for the able-bodied representative subject, the best case achieved a $1.6{^\circ}$ MAE with an $r$ of $0.99$ within the $(1.35, 1.4)$ source-target pairing (Fig. \ref{fig:able_bodied_representative}). The median case yielded a $3.1{^\circ}$ MAE with an $r$ of $0.97$ at the $(0.6, 1.2)$ pairing, while the worst case had an $8.8{^\circ}$ MAE with an $r$ of $0.82$ at the $(0.8, 1.75)$ pairing. For stroke, the best case achieved a $5.0{^\circ}$ MAE with an $r$ of $0.93$ at the $(0.3, 0.6)$ pairing, the median case achieved a $6.1{^\circ}$ MAE with an $r$ of $0.95$ at the $(0.9, 0.3)$ pairing, and the worst case achieved a $13.1{^\circ}$ MAE with an $r$ of $0.64$ at the $(0.6, 0.9)$ pairing (Fig. \ref{fig:stroke_representative}). As a temporal metric to assess adaptation to varied target speeds, the generated able-bodied cadence matched the real cadence with an MAE of 8.5 steps/min ($8.5\%$ mean absolute percentage error across source-target pairings) (Fig. \ref{fig:cadence}). For stroke subjects, the MAE was 8.1 steps/min ($9.0\%$ mean absolute percentage error).

\begin{table}[t]
    \centering
    \caption{Inter-Joint Coordination of Real and Generated Gait}
    \label{tab:interjoint}
    \resizebox{\columnwidth}{!}{%
    \begin{tabular}{lcccccc}
        \toprule
         & \multicolumn{3}{c}{Bilateral Lower Limb}
         & \multicolumn{3}{c}{Intra-Limb} \\
        \cmidrule(lr){2-4}
        \cmidrule(lr){5-7}
         & Real & Generated & $|\Delta R^2|$
         & Real & Generated & $|\Delta R^2|$ \\
        \midrule
        Able-Bodied
            & 0.84 & 0.86 & 0.05
            & 0.38 & 0.43 & 0.13 \\
        \addlinespace[3pt]
        Stroke
            & 0.78 & 0.80 & 0.10
            & & & \\
        \quad Non-Paretic
            & & &
            & 0.50 & 0.56 & 0.19 \\
        \quad Paretic
            & & &
            & 0.45 & 0.55 & 0.22 \\
        \bottomrule
    \end{tabular}%
    }
\end{table}

We further evaluated inter-joint coordination by comparing the multiple $R^2$ between real and generated gait (Table \ref{tab:interjoint}). Excluding same-speed pairs, the generated gait closely matched the real bilateral lower limb coordination, with an $|\Delta R^2|$ of $0.05$ and $0.10$ for able-bodied and stroke subjects, respectively, while the generated intra-limb $R^2$ was slightly higher than the real $R^2$ particularly for the paretic limb.

\subsection{Personalization Rank}

\begin{figure}[t]
    \includegraphics[width=\columnwidth]{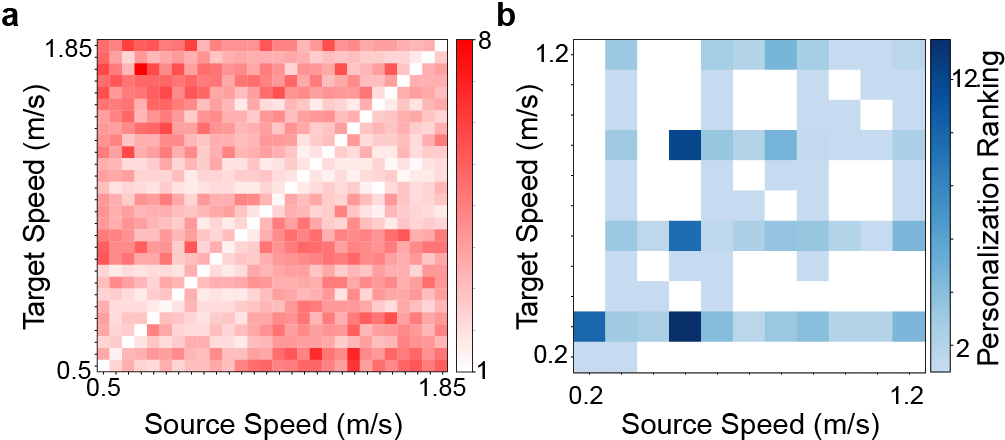}
    \caption{\textbf{Personalization rank heatmap.} Lower ranks indicate stronger subject identity preservation, meaning the generated profile tracks closer to the true subject. Panels show results for (a) able-bodied and (b) stroke cohorts. Stroke speeds are binned to accommodate self-selected speeds, with unrepresented speed pairings shown as white cells.}
    \label{fig:personalization_heatmap}
\end{figure}

The average personalization rank for each source-target speed pairing across subjects is shown in the heatmap (Fig. \ref{fig:personalization_heatmap}). The maximum able-bodied rank was 16 for a subject with source-target pairing of $(0.6, 1.6)$ and 16 for a stroke subject with the pairing of $(0.47, 0.3)$. 

\subsection{Speed Analysis}

\begin{figure}[t]
    \includegraphics[width=\columnwidth]{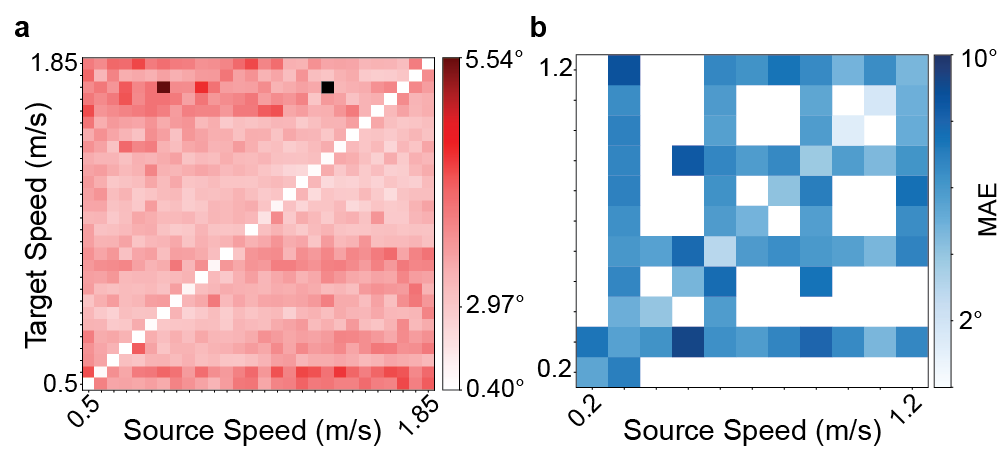}
    \caption{\textbf{Error heatmap.} Distribution of MAE across speed pairings. (a) Error heatmap for able-bodied subjects, plotted with outliers omitted to improve readability. (b) Error heatmap for stroke subjects. Stroke speeds are binned to accommodate self-selected speeds, with unrepresented speed pairings shown as white cells.}
    \label{fig:heatmaps}
\end{figure}

\begin{figure}[t]
    \includegraphics[width=\columnwidth]{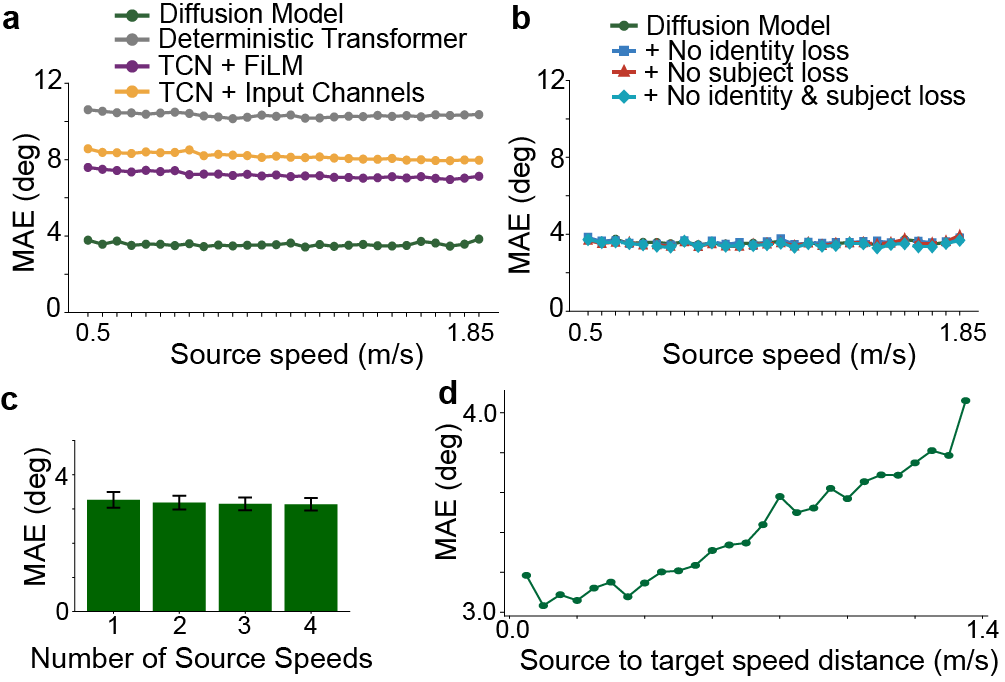}
    \caption{\textbf{Performance under various source speed conditions.} (a) MAE per source speed across the diffusion model, deterministic transformer, and TCN baselines. (b) MAE per source speed for the diffusion model and auxiliary loss ablations. (c) Source-speed availability analysis on able-bodied data. For each subset size, the nearest available source speed was selected independently for each target speed, and only that source sequence was used for generation. Bars show the mean and standard deviation across subjects. (d) Average able-bodied MAE as a function of the distance between source and target speeds.}
    \label{fig:speed_analysis}
\end{figure}

The errors were low across most non-identity speed pairings in both datasets (Fig. \ref{fig:heatmaps}). Averaged across subjects, the able-bodied MAE remained below $5^\circ$ at every source-target distance and increased with distance (Fig. \ref{fig:speed_analysis}d), while the average MAE per source speed varied by approximately $0.4^\circ$. Individual pairings could exceed this, with a maximum able-bodied error of $23.4{^\circ}$ on the $(1.45, 1.75)$ source-target pairing. For stroke subjects, the maximum error was $28.1{^\circ}$ on the $(0.9, 0.3)$ pairing, a $4.7{^\circ}$ increase over the able-bodied worst case.

We also analyzed how the number of available source speeds affects generation accuracy (Fig. \ref{fig:speed_analysis}c). For each number of available source speeds, we considered all possible subsets of that size. For each subset, the model used only one source gait sequence for each target speed not included in the subset, selected from the nearest available source speed, with ties assigned to the higher speed. We then averaged the MAE across the evaluated target speeds and subsets for each subject, followed by averaging across subjects. This analysis evaluates the effect of source-speed availability rather than jointly conditioning the model on multiple source sequences. The average MAE was $3.27^\circ$ with a single source speed and $3.14^\circ$ with four source speeds, a difference of $0.13^\circ$ (Fig. \ref{fig:speed_analysis}c).

To test if personalization relies on the correct subject's gait, we replaced the source gait with that of a randomly selected different subject at the same source speed. For stroke subjects, the analysis was restricted to the fixed walking speeds (0.3, 0.6, 0.9, and 1.2 m/s). The subject embedding, denoiser input, and reconstruction all used the other subject's gait, and we evaluated rank against the originally intended subject. This increased the personalization rank from 3.3 to 14.4 for able-bodied and 3.0 to 6.5 for stroke subjects (Table \ref{tab:randomized}).

\begin{table}[!t]
    \centering
    \caption{Personalization Rank with Correct and Randomly Reassigned Source Gait}
    \label{tab:randomized}
    \begin{tabular}{lcc}
    \toprule
     & Correct Source Gait & Reassigned Source Gait \\
    \midrule
    Able-Bodied & 3.3 & 14.4 \\
    Stroke & 3.0 & 6.5 \\
    \bottomrule
    \end{tabular}
\end{table}

\subsection{Baseline Models}
We evaluated the proposed diffusion model against deterministic transformer and TCN baselines, along with the ablations of the auxiliary losses (Figs. \ref{fig:speed_analysis}a and \ref{fig:speed_analysis}b). Under the matched 30-epoch training setting, the full diffusion model achieved an able-bodied MAE of $3.6^{\circ}$ compared to $10.3^{\circ}$ for the deterministic transformer with the same backbone and conditioning mechanism, demonstrating the benefit of iterative diffusion-based generation over direct residual prediction in this setting. The TCN with FiLM conditioning had an MAE of $7.2^{\circ}$, compared to $8.2^{\circ}$ with input-channel conditioning, suggesting that modulating the intermediate representations was more effective than directly incorporating conditioning information into the input. Removing the identity preservation loss, subject preservation loss, or both resulted in able-bodied MAEs of $3.6^{\circ}$, $3.5^{\circ}$, and $3.5^{\circ}$, respectively, compared to $3.6^{\circ}$ for the full model trained for the same 30 epochs. However, the full model achieved the lowest stroke MAE of $6.3^{\circ}$, compared to $6.9^{\circ}$, $7.3^{\circ}$, and $6.9^{\circ}$ for the respective ablations.

\section{Discussion}
\subsection{Key Findings}
The results support the feasibility of generating subject-specific target-speed gait kinematics from a single source-speed sequence under the evaluated conditions (Table \ref{tab:overallMAE}). Consistent with prior studies \cite{kneeAngle, Li2024}, knee joints showed a higher error than the hip and ankle. Low personalization ranks for both able-bodied and stroke subjects (Table \ref{tab:overallMAE}) show that the model preserved subject identity rather than producing population-average gait. The model also preserved bilateral lower-limb inter-joint coordination (Table \ref{tab:interjoint}), consistent with FiLM modulating a representation shared across all joints.

Since stroke speeds (0.3-1.2 m/s) extended below the able-bodied training range (0.5-1.85 m/s), stroke cadence analysis was restricted to target speeds within the overlapping range of 0.5-1.2 m/s for a fair comparison. With this, the generated cadence followed the real increasing trend with target speed for both able-bodied (8.5\% error) and stroke (9.0\% error) subjects, indicating that the model captures the temporal scaling of gait.

\subsection{Generalization to Stroke Gait}
The model generalized to stroke data, unseen during training, across MAE, personalization, and cadence, without stroke-specific normalization or fine-tuning. MAE was lower on the paretic side than the non-paretic side overall and across joints. This difference may partly reflect smaller joint ranges of motion, since absolute error depends on movement amplitude. Representative trajectories illustrate the generated joint-angle profiles for selected able-bodied and stroke subjects (Figs. \ref{fig:able_bodied_representative} and \ref{fig:stroke_representative}). The inter-joint coordination analysis indicates similar bilateral lower-limb linear relationships in real and generated gait for both populations. However, generated gait showed higher mean intra-limb $R^2$ than real gait, particularly on the paretic side (0.55 versus 0.45), indicating overestimation of linear coupling within that limb. Our evaluation was limited to sagittal-plane joint kinematics and therefore did not assess frontal- or transverse-plane deviations such as hip circumduction. Evaluating these movements would require extending the framework to include those planes.

\subsection{Speed Analysis and Deployment}
Errors generally increased with larger source-target speed differences, although not consistently across all pairings (Fig. \ref{fig:heatmaps}). Average error was nearly uniform across source speeds ($0.4^\circ$ range), so no single source speed clearly outperformed the others. Moreover, adding source speeds reduced MAE by only $2.5\%$ from one to two source speeds and $0.13^\circ$ from one to four (Fig. \ref{fig:speed_analysis}c). Under the evaluated source-selection protocol, access to additional source speeds provided limited improvement in average accuracy, suggesting potential to reduce data collection requirements using a single source-speed recording.

\subsection{Personalization Rank}
The personalization rank showed a trend similar to MAE, where higher ranks occurred at extreme source-target speed differences (Fig. \ref{fig:personalization_heatmap}). The low rank values extended beyond the diagonal, indicating that the model preserved subject-specific gait features even across non-zero speed gaps. However, as seen in the maximum error cases, the worst rank cases were not limited to the most extreme pairings, suggesting that personalization depends on subject-specific gait variation and inter-subject similarity at specific target speeds. The higher personalization rank with randomly reassigned source gait (Table \ref{tab:randomized}) supports the importance of using the intended subject's source trajectory for personalization.

\subsection{Baseline and Ablation Analysis}
Under the matched training protocol, diffusion outperformed the deterministic transformer with the same backbone and conditioning, supporting iterative denoising for this task. The TCN baselines also achieved higher errors, suggesting that the attention-based backbone combined with diffusion was more effective for this task than the evaluated convolutional alternatives. Additionally, the TCN with FiLM conditioning achieved lower error than the input-channel conditioning version, supporting the effectiveness of FiLM for incorporating subject and speed information. The auxiliary loss ablations showed similar or slightly lower MAEs for able-bodied subjects when the identity and subject preservation losses were removed. However, removing these losses increased the MAE for stroke subjects, suggesting that the auxiliary losses may contribute to generalization to unseen clinical gait patterns.

\subsection{Limitations}
The stroke evaluation was limited by the number of stroke subjects and available speed pairings. Larger and more diverse clinical datasets are needed to assess generalization more broadly. Similarly, the personalization rank metric was dependent on the availability of comparable subjects at the target speed. Additionally, our evaluation was limited to sagittal-plane joint kinematics, cadence, and linear inter-joint relationships, which do not fully capture the biomechanical plausibility of the generated gait. Future work will incorporate additional biomechanical constraints and evaluations to assess the physical consistency of generated motion. 

Gait-cycle-averaged MAE also does not capture cycle-to-cycle variability or temporal errors, which analysis of the raw sequences could assess. Because source and target gait events are not aligned, applications requiring synchronized kinematic and kinetic data would need gait-event landmarks. Although we have not directly tested how our pipeline works in downstream tasks, potential use cases include imitation-learned user-specific digital twins in physics-informed simulations for training exoskeleton control policies \cite{Choi2026, chiu2025learning, Leem2026} and kinematics-based lower-limb prosthetic controllers during swing phase \cite{Tran2022, Quintero2018}.

\subsection{Future Work}
Future work will scale to a larger number of stroke subjects with larger inter-subject variability. We will also synthesize a wider range of daily locomotor tasks such as ramp or stair walking. Future work will also extend the framework to frontal- and transverse-plane kinematics to better capture gait deviations such as hip circumduction in stroke patients. We will also extend this pipeline to estimate user-specific kinetic and muscle activation profiles for exoskeleton control and biomechanical monitoring. Additionally, computer vision-based estimation could also provide source kinematics without specialized motion capture equipment \cite{Gilon2026, Song2025, Kanko2021}.

\section{Conclusion}
This work introduced a subject-conditioned residual diffusion framework that generates personalized lower-limb kinematics at target walking speeds from a single source-speed sequence. It achieved a $3.27^\circ$ MAE on unseen able-bodied subjects, generalized to stroke subjects without fine-tuning, and outperformed deterministic transformer and TCN baselines. These results suggest generating target-speed gait from a single source-speed recording could reduce the motion capture data needed for personalized gait modeling, with potential applications in wearable and rehabilitation robotics.

\bibliographystyle{ieeetr}
\bibliography{references}

\end{document}